\newif\ifdebug
\debugfalse

\documentclass[
twocolumn,
]{TEMPLATE/ceurart}

\usepackage{listings}
\usepackage{listings}
\usepackage{xcolor}

\usepackage{svg}
\usepackage{pgfplots}

\usepackage{tikz}
\usetikzlibrary{positioning, shapes, arrows.meta}

\usepackage[nameinlink]{cleveref}

\usepackage{todonotes}
\usepackage{marginnote}

\usepackage{colortbl}
\usepackage{multirow}
\usepackage{multicol}
\usepackage{enumitem}
\usepackage{dsfont}
\usepackage{lipsum}
\usepackage{caption}
\usepackage{placeins}

\definecolor{anne}{rgb}{0,0.5,0.9}
\definecolor{valentin}{rgb}{0.998,0.722,0.635}
\definecolor{anna}{rgb}{0,0.5,0.9}
\definecolor{simon}{rgb}{0.998,0.722,0.635}

\definecolor{darkgreen}{rgb}{0.0, 0.5, 0.0}

\ifdebug
    \paperwidth=\dimexpr \paperwidth + 20cm\relax
    \oddsidemargin=\dimexpr\oddsidemargin + 10cm\relax
    \evensidemargin=\dimexpr\evensidemargin + 10cm\relax
    \marginparwidth=\dimexpr \marginparwidth + 10cm - \marginparsep\relax

    \newcommand{\valentintodo}[2][valentin]{\todo[color=#1,size=\footnotesize]{\textbf{VK:} #2}}
    \newcommand{\annetodo}[2][anne!30]{\todo[color=#1,size=\footnotesize]{\textbf{AF:} #2}}
    \newcommand{\annatodo}[2][anna]{\todo[color=#1,size=\footnotesize]{\textbf{AH:} #2}}
    \newcommand{\simontodo}[2][simon!30]{\todo[color=#1,size=\footnotesize]{\textbf{SR:} #2}}

    \newcommand{\valentintodofigure}[2][0cm]{\marginnote{\todo[color=valentin,size=\footnotesize,inline]{\textbf{VK:} #2}}[#1] }
    \newcommand{\annetodofigure}[2][0cm]{\marginnote{\todo[color=anne!30,size=\footnotesize,inline]{\textbf{AF:} #2}}[#1] }
    \newcommand{\annatodofigure}[2][0cm]{\marginnote{\todo[color=anna,size=\footnotesize,inline]{\textbf{AH:} #2}}[#1] }
    \newcommand{\simontodofigure}[2][0cm]{\marginnote{\todo[color=simon!30,size=\footnotesize,inline]{\textbf{SR:} #2}}[#1] }

\else
    \newcommand{\valentintodo}[2][valentin]{}
    \newcommand{\annetodo}[2][anne!30]{}
    \newcommand{\annatodo}[2][anna]{}
    \newcommand{\simontodo}[2][simon!30]{}

    \newcommand{\valentintodofigure}[2][0cm]{}
    \newcommand{\annetodofigure}[2][anne!30]{}
    \newcommand{\annatodofigure}[2][anna]{}
    \newcommand{\simontodofigure}[2][simon!30]{}

\fi

\pgfplotsset{compat=1.18}

\definecolor{codegreen}{rgb}{0,0.6,0}
\definecolor{codegray}{rgb}{0.5,0.5,0.5}
\definecolor{codepurple}{rgb}{0.58,0,0.82}
\definecolor{backcolour}{rgb}{0.95,0.95,0.92}

\definecolor{turq}{HTML}{00ffd9}
\definecolor{highlightblue}{HTML}{00c5ff}

\newcommand{\highlight}[2]{%
  {%
    \setlength{\fboxsep}{1pt}%
    \colorbox{#1}{#2}%
  }%
}%

\usepackage{amssymb}%
\usepackage{pifont}%
\newcommand{\yes}{\textcolor{green}{\ding{51}}}%
\newcommand{\no}{\textcolor{red}{\ding{55}}}%

\lstdefinestyle{jsonstyle}{
    backgroundcolor=\color{backcolour},
    commentstyle=\color{codegreen},
    keywordstyle=\color{magenta},
    numberstyle=\tiny\color{codegray},
    stringstyle=\color{codepurple},
    basicstyle=\ttfamily\footnotesize,
    breakatwhitespace=false,
    breaklines=true,
    captionpos=b,
    keepspaces=true,
    numbers=left,
    numbersep=5pt,
    showspaces=false,
    showstringspaces=false,
    showtabs=false,
    tabsize=2
}

\begin{document}

\copyrightyear{2022}
\copyrightclause{Copyright for this paper by its authors.
    Use permitted under Creative Commons License Attribution 4.0
    International (CC BY 4.0).}

\conference{6th Workshop on Patent Text Mining and Semantic Technologies (PatentSemTech) 2025}

\def\DatasetName{PEDANTIC\xspace}
\title{PEDANTIC: A Dataset for the Automatic Examination\\of Definiteness in Patent Claims}

\author[1,2]{Valentin Knappich}[%
    email=valentin.knappich@de.bosch.com
]
\address[1]{Bosch Center for AI}
\address[2]{University of Augsburg}

\author[2]{Annemarie Friedrich}[%
    email=annemarie.friedrich@uni-a.de
]

\author[1]{Anna Hätty}[%
    email=anna.haetty@de.bosch.com
]

\author[3]{Simon Razniewski}[%
    email=simon.razniewski@tu-dresden.de
]
\address[3]{ScaDS.AI \& TU Dresden}

\definecolor{delim}{RGB}{20,105,176}
\definecolor{numb}{RGB}{106, 109, 32}
\definecolor{string}{rgb}{0.64,0.08,0.08}

\lstdefinelanguage{json}{
    numbers=left,
    numberstyle=\small,
    frame=single,
    rulecolor=\color{black},
    showspaces=false,
    showtabs=false,
    breaklines=true,
    postbreak=\raisebox{0ex}[0ex][0ex]{\ensuremath{\color{gray}\hookrightarrow\space}},
    breakatwhitespace=true,
    basicstyle=\ttfamily\small,
    upquote=true,
    morestring=[b]",
    stringstyle=\color{string},
    literate=
    *{0}{{{\color{numb}0}}}{1}
    {1}{{{\color{numb}1}}}{1}
    {2}{{{\color{numb}2}}}{1}
    {3}{{{\color{numb}3}}}{1}
    {4}{{{\color{numb}4}}}{1}
    {5}{{{\color{numb}5}}}{1}
    {6}{{{\color{numb}6}}}{1}
    {7}{{{\color{numb}7}}}{1}
    {8}{{{\color{numb}8}}}{1}
    {9}{{{\color{numb}9}}}{1}
    {\{}{{{\color{delim}{\{}}}}{1}
    {\}}{{{\color{delim}{\}}}}}{1}
    {[}{{{\color{delim}{[}}}}{1}
    {]}{{{\color{delim}{]}}}}{1},
}

\cortext[1]{Corresponding author.}
\fntext[1]{These authors contributed equally.}

\begin{abstract}
    Patent claims define the scope of protection for an invention.
    If there are ambiguities in a claim, it is rejected by the patent office.
    In the US, this is referred to as indefiniteness (35 U.S.C § 112(b)) and is among the most frequent reasons for patent application rejection.
    The development of automatic methods for patent definiteness examination has the potential to make patent drafting and examination more efficient, but no annotated dataset has been published to date.

    We introduce
    \DatasetName (\underline{P}at\underline{e}nt \underline{D}efiniteness Ex\underline{a}mi\underline{n}a\underline{ti}on \underline{C}orpus),
    a novel dataset of 14k US patent claims from patent applications relating to Natural Language Processing (NLP), annotated with reasons for indefiniteness.
    We construct \DatasetName using a fully automatic pipeline that retrieves office action documents from the USPTO and uses Large Language Models (LLMs) to extract the reasons for indefiniteness.
    A human validation study confirms the pipeline's accuracy in generating high-quality annotations.
    To gain insight beyond binary classification metrics, we implement an LLM-as-Judge evaluation that compares the free-form reasoning of every model-cited reason with every examiner-cited reason.
    We show that LLM agents based on Qwen 2.5 32B and 72B struggle to outperform logistic regression baselines on definiteness prediction, even though they often correctly identify the underlying reasons.
    \DatasetName provides a valuable resource for patent AI researchers, enabling the development of advanced examination models.
    We release the dataset and code at \url{https://github.com/boschresearch/pedantic-patentsemtech}.
\end{abstract}


\maketitle

\section{Introduction}\label{sec:intro}

The patent system plays a crucial role in fostering innovation by granting inventors exclusive rights to their inventions.
Central to this system are patent claims, which are concise and precise statements that define the metes and bounds of the protected invention.
The process of obtaining a patent involves rigorous examination by patent offices to ensure that the application meets specific criteria, including novelty, non-obviousness, and, critically, \textit{definiteness}.
The latter requires that every claim is sufficiently clear and unambiguous to enable a person skilled in the art (called \textit{the Person of Ordinary Skill in the Art}, or \textit{POSITA}) to understand the scope of the invention.
In American patent law, this is defined in 35 U.S.C. § 112(b) 
(comparable to \textit{clarity} in the EU\footnote{\url{https://www.epo.org/en/legal/epc/2020/a84.html}\\\url{https://www.epo.org/en/legal/guidelines-epc/2024/f_iv_4.html}}), which states that \enquote{\textit{the specification shall conclude with one or more claims particularly pointing out and distinctly claiming the subject matter which the applicant regards as his invention.}}
The Manual of Patent Examining Procedure (MPEP) \cite{usptoManualPatentExamining} provides detailed instructions for the examination.
We present the most common categories of indefiniteness in \Cref{tab:categories}.

Ensuring definiteness is challenging for patent attorneys and examiners.
Patent applications typically undergo multiple rejection-response cycles, averaging 26 months from filing to disposition \cite{usptoPendencyPatentsDashboard}.
With increasing application volumes \cite{worldintellectualpropertyorganization.WorldIntellectualProperty00}, AI-powered methods are needed to improve examination efficiency and consistency.
Automating examination could yield significant cost savings. 
For instance, it could assist examiners, aid attorneys in drafting more robust applications, and improve accessibility for those without extensive patent expertise.
It could also provide feedback for training and evaluating automatic drafting systems \cite{knappich2024pap2pat}.

Indefiniteness is one of the most common reasons for rejection \cite{luUSPTOPatentProsecution2017,MostCommonRejections2019,loPretrainedTransformerbasedClassification2021}, yet its automatic examination has received very little attention in prior work compared to novelty and non-obviousness.
To bridge this gap, we specifically target the automatic examination of definiteness.
We argue that predicting definiteness as a binary label is insufficient for practical applications.
For patent drafters and examiners, knowing that a claim is indefinite does not provide actionable information.
Rather, understanding \textit{why} it is indefinite is crucial, as it enables targeted improvements to the claim.
For this purpose, we introduce \DatasetName, a dataset of patent claims annotated with detailed justifications for rejection due to indefiniteness.
In addition to the claims, \DatasetName includes fine-grained indefiniteness categories (such as \textit{'antecedent basis'}) along with a free-form reasoning, and the affected ranges of the claim for every indefiniteness reason.
We create it fully automatically, leveraging Large Language Models (LLMs) to parse USPTO office actions into a structured format that includes these annotations.
In total, \DatasetName includes 14k claims from 3k utility patent applications relating to Natural Language Processing (NLP) filed after 2014.
It facilitates a nuanced evaluation, distinguishing between systems that identify indefiniteness based on superficial clues and those that are able to pinpoint the correct underlying issue.
In addition to established classification metrics, we implement a reference-based LLM-as-Judge \cite{zheng2023judge} evaluation that compares each model-cited rejection reason with every examiner-cited rejection reason.
We release our dataset and code publicly to facilitate further research.

\begin{figure*}
    \resizebox{\textwidth}{!}{
        \includesvg{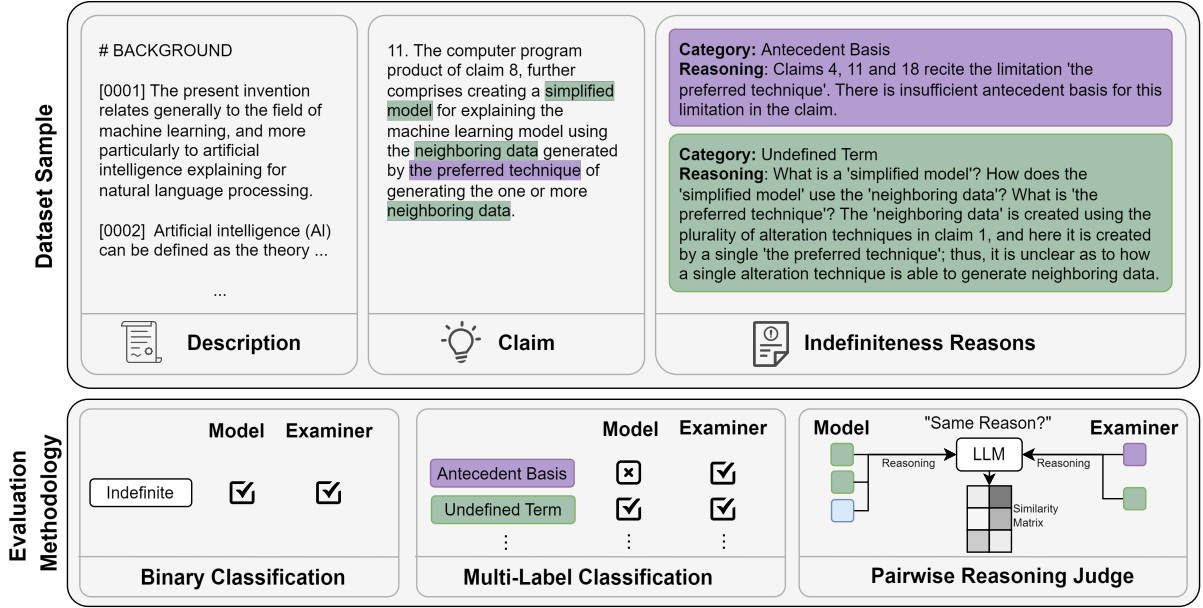}
    }
    \caption{
        \DatasetName example and evaluation methodology.
        Every sample includes a claim, the patent description, and a list of rejection reasons that cause indefiniteness, each with a free-form reasoning, indefiniteness category, and recited claim spans.
        We evaluate models on (1) the prediction of the binary label definite or indefinite, (2) the identification of the correct indefiniteness categories as multi-label classification, and (3) the generation of correct free-form reasoning for indefiniteness reasons using an LLM-as-Judge that compares all pairs of examiner-cited reason and model-cited reason.
    }
    \label{fig:example}
\end{figure*}

\section{Related Work}\label{sec:related_work}

\paragraph{Automated Patent Examination.}
Most related work on automatic patent examination has focused on novelty and non-obviousness.
Datasets for novelty assessment 
\cite{rischPatentMatchDatasetMatching2020, vowinckelSEARCHFORMERSemanticPatent2023, DBLP:conf/patentsemtech/BlumeHH24, DBLP:conf/patentsemtech/ParikhD24} 
pair patent claims either with short passages 
\cite{rischPatentMatchDatasetMatching2020, vowinckelSEARCHFORMERSemanticPatent2023} 
or large chunks 
\cite{DBLP:conf/patentsemtech/BlumeHH24, DBLP:conf/patentsemtech/ParikhD24} 
from prior art.
They use citations marked as novelty-destroying by the examiner as positive samples and obtain negative samples either from other citations 
\cite{rischPatentMatchDatasetMatching2020, vowinckelSEARCHFORMERSemanticPatent2023, DBLP:conf/patentsemtech/BlumeHH24} 
and/or from related patents
\cite{vowinckelSEARCHFORMERSemanticPatent2023, DBLP:conf/patentsemtech/ParikhD24}.
Novelty has been evaluated using BERT-based models
\cite{rischPatentMatchDatasetMatching2020,vowinckelSEARCHFORMERSemanticPatent2023,DBLP:conf/patentsemtech/BlumeHH24,DBLP:conf/patentsemtech/ParikhD24,stamatisNovelRerankingArchitecture2024,shanStructuralRepresentationLearning2024},
graph neural networks 
\cite{shenPatentGrapherPLMGNNsHybrid2024, weiExtractionNoveltyEvaluation2024},
and LLMs
\cite{DBLP:conf/patentsemtech/BlumeHH24, ikomaCanAIExamine2025}.
\citet{leePatentEditsFramingPatent2024} furthermore present a dataset for the prediction of patent edits following rejections based on novelty and non-obviousness.

\paragraph{Patent Clarity and Definiteness.}
\citet{hidoModelingPatentQuality2012} predict patentability using linguistic features including syntactic complexity and word age.
\citet{kongLinguisticMetricsPatent2023} model patent readability using § 112(a) (lack of disclosure) with linguistic features, finding university-issued patents more readable than corporate-issued ones.
\citet{loPretrainedTransformerbasedClassification2021} frame patentability prediction as a multi-label classification task, including definiteness, using BERT-based models, but do not report definiteness-specific results.
\citet{ashtorModelingPatentClarity2022} trains a definiteness classifier based on linguistic features and reaches 68\% AUROC.
They use this classifier as proxy for clarity and show that clarity has improved over time through policy changes regarding definiteness rejections.

\paragraph{LLM Explanations.}
While the primary purpose of a classifier is to predict the most likely class, practical applications often demand more transparency and insight into the reason behind a prediction.
There is a large body of work concerning explainable AI (XAI) \cite{MERSHA2024128111} that has developed various techniques to produce explanations for a classifier's prediction, including feature attribution methods, counterfactual explanations, rule extraction, and example-based reasoning.
Recently, LLMs have been employed to directly generate explanations along with their predictions.
Such self-explanations have been shown to perform on par with traditional explainability methods \cite{huang2023largelanguagemodelsexplain}, but also have limited faithfulness \cite{madsenAreSelfexplanationsLarge2024, agarwal2024faithfulnessvsplausibilityunreliability}.
In this work, we propose to use the free-form reasons written by the patent examiner in the rejection full-text as ground truth to evaluate whether a model predicts indefiniteness for the right reasons.

\begin{table*}[t]
    \centering

    \begin{tabular}{lcl}
        \toprule
        Reason                       & MPEP Section & Description                                          \\
        \midrule

        Antecedent Basis             &
        2173.05(e)                   &
        \makecell[l]{Claim contains a term referencing an element lacking a clear prior introduction,      \\creating ambiguity what it references.} \\
        \cmidrule[0.01pt]{1-3}

        Undefined Term               &
        2173.05(a)                   &
        \makecell[l]{A term lacks a clear, accepted and/or unambiguous meaning to a POSITA,                \\making the claim's scope uncertain.} \\
        \cmidrule[0.01pt]{1-3}

        Relative Term                &
        2173.05(b)                   &
        \makecell[l]{A relative term (e.g., 'thin,' 'substantial') is used without providing a clear point \\of comparison, rendering the claim's scope indefinite.} \\
        \cmidrule[0.01pt]{1-3}

        Exemplary Phrasing           &
        2173.05(d)                   &
        \makecell[l]{Claim uses 'such as,' or similar phrasing, making it unclear whether the              \\listed items are exhaustive or merely examples, leading to indefiniteness.} \\
        \cmidrule[0.01pt]{1-3}

        Functional Claiming          &
        2173.05(g)                   &
        \makecell[l]{Claim recites 'means for' or 'step for' without disclosing adequate                   \\corresponding structure, material, or acts in the specification, as required\\ under 35 U.S.C. 112(f) or pre-AIA equivalent.} \\
        \cmidrule[0.01pt]{1-3}

        Contradicting Limitations    &
        2173                         &
        \makecell[l]{Claim includes an element that contradicts or is inconsistent with other              \\claim limitations, making the claim's scope unclear.} \\
        \cmidrule[0.01pt]{1-3}

        \makecell[l]{Omission of Essential                                                                 \\Elements or Steps}  &
        2172.01                      &
        \makecell[l]{Claim fails to recite an element, step, or cooperative relationship between           \\elements/steps that is essential to the invention as disclosed.} \\
        \cmidrule[0.01pt]{1-3}

        \arrayrulecolor{gray}
        \textcolor{gray}{Dependence} &
        \textcolor{gray}{-}          &
        \textcolor{gray}{\makecell[l]{Claim depends on an indefinite claim.}}                              \\
        \cmidrule[0.01pt]{1-3}

        \textcolor{gray}{Other}      &
        \textcolor{gray}{-}          &
        \textcolor{gray}{\makecell[l]{Catch-all category for indefiniteness reasons not covered above.}}   \\

        \arrayrulecolor{black}
        \bottomrule
    \end{tabular}
    \caption{Brief descriptions of indefiniteness categories and their corresponding MPEP sections. The same descriptions were used during parsing to help the model identify the correct reason. Greyed-out reasons were used only during parsing and filtered out for the final dataset.}
    \label{tab:categories}
\end{table*}

\section{\DatasetName Dataset}\label{sec:dataset}

In this section, we describe the creation of our dataset, including the retrieval of seed patent applications, the retrieval of rejection notices, document parsing, and dataset splitting.

\subsection{Seed Patent Applications}\label{subsec:seed_patents}

Our dataset creation pipeline first requires a set of seed patent applications.
We focus on patent applications in the area of NLP, i.e., applications with the CPC class \enquote{\textit{G06F40}} (\enquote{Handling natural language data}), but our pipeline is agnostic to this and can be readily used with seed applications from any other field.
We query the USPTO Open Data Portal (ODP)\footnote{\url{https://data.uspto.gov/}} API for applications from 2014 to date with at least one filed rejection notice.
We do not consider applications prior to 2014 because the requirements for definiteness changed significantly after the US Supreme Court's ruling on Nautilus vs. Biosig \cite{NautilusIncBiosig}.
Before 2014, a claim was only rejected for indefiniteness if it was \enquote{insolubly ambiguous}, whereas afterward, indefiniteness was interpreted as a claim \enquote{[failing] to inform, with reasonable certainty, those  skilled in the art about the scope of the invention}.
\citet{ashtorModelingPatentClarity2022} finds empirical evidence that this ruling has significantly increased the fraction of claims being rejected for indefiniteness and the average claim clarity.
To avoid exploitable biases, we filter out applications filed before 2014.

\subsection{Document Download}\label{subsec:download_and_parse}

For all retrieved seed patents, we download the claim, specification and office action documents from the ODP File Wrapper API\footnote{\url{https://data.uspto.gov/apis/patent-file-wrapper/search}}.
We only consider the first office action, as it generally establishes the core grounds for rejection.
While subsequent office actions may address remaining or newly introduced issues, the initial rejection often provides the most comprehensive overview of the examiner's concerns.
We select the latest claim and specification documents preceding this office action and download all three documents in XML format.
We parse the XML documents into Markdown and JSON for easier further processing.

\begin{table*}
    \valentintodofigure{Is it confusing that the percentages sum up to 100\% per row for the total sections but per column for the other sections?}
    \centering
    \begin{tabular}{cccccc}
    \toprule
                                  &                                 & \multicolumn{4}{c}{Dataset Splits}                                                                                                                       \\
    \cmidrule(lr){3-6}
                                  &                                 & All                                 & Train                                 & Test                                 & Validation                          \\
    \midrule
    \multirow{5}{*}{Claims}       & Total                           & \phantom{}14536 \phantom{}(100.00\%)                      & \phantom{0}8730 \phantom{0}(60.06\%)                      & \phantom{0}4362 \phantom{0}(30.01\%)                      & \phantom{0}1444 \phantom{00}(9.93\%)                      \\
    \cmidrule(lr){2-6}
                                  & Definite                        & \phantom{0}7268 \phantom{0}(50.00\%)                        & \phantom{0}4363 \phantom{0}(49.98\%)                        & \phantom{0}2178 \phantom{0}(49.93\%)                        & \phantom{00}727 \phantom{0}(50.35\%)                        \\
                                  & Indefinite                      & \phantom{0}7268 \phantom{0}(50.00\%)                      & \phantom{0}4367 \phantom{0}(50.02\%)                      & \phantom{0}2184 \phantom{0}(50.07\%)                      & \phantom{00}717 \phantom{0}(49.65\%)                      \\
    \cmidrule(lr){2-6}
                                  & Independent                     & \phantom{0}3339 \phantom{0}(22.97\%)                        & \phantom{0}1996 \phantom{0}(22.86\%)                        & \phantom{0}1026 \phantom{0}(23.52\%)                        & \phantom{00}317 \phantom{0}(21.95\%)                        \\
                                  & Dependent                       & \phantom{}11197 \phantom{0}(77.03\%)                        & \phantom{0}6734 \phantom{0}(77.14\%)                        & \phantom{0}3336 \phantom{0}(76.48\%)                        & \phantom{0}1127 \phantom{0}(78.05\%)                        \\
    \midrule
    \multirow{3}{*}{Applications} & Total                           & \phantom{0}3710 \phantom{}(100.00\%)                    & \phantom{0}2226 \phantom{0}(60.00\%)                    & \phantom{0}1113 \phantom{0}(30.00\%)                    & \phantom{00}371 \phantom{0}(10.00\%)                    \\
    \cmidrule(lr){2-6}
                                  & Definite                        & \phantom{0}1855 \phantom{0}(50.00\%)                      & \phantom{0}1107 \phantom{0}(49.73\%)                      & \phantom{00}559 \phantom{0}(50.22\%)                      & \phantom{00}189 \phantom{0}(50.94\%)                      \\
                                  & Indefinite                      & \phantom{0}1855 \phantom{0}(50.00\%)                    & \phantom{0}1119 \phantom{0}(50.27\%)                    & \phantom{00}554 \phantom{0}(49.78\%)                    & \phantom{00}182 \phantom{0}(49.06\%)                    \\
    \midrule
    \multirow{9}{*}{\makecell{Rejection                                                                                                                                                                                        \\Reason}} & Total       & \phantom{0}9215 \phantom{}(100.00\%)    & \phantom{0}5598 \phantom{0}(60.75\%)    & \phantom{0}2765 \phantom{0}(30.01\%)    & \phantom{00}852 \phantom{00}(9.25\%)    \\
    \cmidrule(lr){2-6}
                                  & Antecedent Basis                & \phantom{0}3350 \phantom{0}(36.35\%)          & \phantom{0}1994 \phantom{0}(35.62\%)          & \phantom{0}1015 \phantom{0}(36.71\%)          & \phantom{00}341 \phantom{0}(40.02\%)          \\
                                  & Undefined Term                  & \phantom{0}3394 \phantom{0}(36.83\%)            & \phantom{0}2093 \phantom{0}(37.39\%)            & \phantom{0}1035 \phantom{0}(37.43\%)            & \phantom{00}266 \phantom{0}(31.22\%)            \\
                                  & Relative Term                   & \phantom{00}910 \phantom{00}(9.88\%)             & \phantom{00}568 \phantom{0}(10.15\%)             & \phantom{00}258 \phantom{00}(9.33\%)             & \phantom{000}84 \phantom{00}(9.86\%)             \\
                                  & Exemplary Phrasing              & \phantom{00}121 \phantom{00}(1.31\%)        & \phantom{000}97 \phantom{00}(1.73\%)        & \phantom{000}19 \phantom{00}(0.69\%)        & \phantom{0000}5 \phantom{00}(0.59\%)        \\
                                  & Functional Claiming             & \phantom{00}850 \phantom{00}(9.22\%)       & \phantom{00}488 \phantom{00}(8.72\%)       & \phantom{00}270 \phantom{00}(9.76\%)       & \phantom{000}92 \phantom{0}(10.80\%)       \\
                                  & Contradicting Limitation        & \phantom{00}442 \phantom{00}(4.80\%) & \phantom{00}288 \phantom{00}(5.14\%) & \phantom{00}113 \phantom{00}(4.09\%) & \phantom{000}41 \phantom{00}(4.81\%) \\
                                  & \makecell{Omission of Essential                                                                                                                                                            \\Elements or Steps} & \phantom{00}148 \phantom{00}(1.61\%)  & \phantom{000}70 \phantom{00}(1.25\%)  & \phantom{000}55 \phantom{00}(1.99\%)  & \phantom{000}23 \phantom{00}(2.70\%) \\
    \bottomrule
\end{tabular}
    \caption{Dataset statistics across splits. In \enquote{Total} sections, the percentages sum to 100\% per row, otherwise per column.}
    \label{tab:stats}
\end{table*}

\subsection{Office Action Parsing}\label{subsec:office_action_parsing}

We parse the office actions' full-text into a structured representation using Gemma 3 27B \cite{gemmateam2025gemma3technicalreport}.
First, we select the sections related to indefiniteness by filtering for sections with headings containing \enquote{112}.
Next, we prompt the LLM with the selected sections and instruct it to extract indefiniteness reasons in a JSON schema.
In particular, each rejection  contains the text snippet arguing why the claim is indefinite, a category from \Cref{tab:categories}, and a list of recited phrases.
We instruct the LLM to extract the free-form argumentation and recited phrases verbatim and not to paraphrase, extend or explain them.
The prompt is attached in \Cref{sec:appendix_parsing_prompt}.
Lastly, we use fuzzy matching to find the occurrences of recited phrases in the claim text.
An example from our dataset is shown in \Cref{fig:example}.
A claim can have multiple rejection reasons, but most have only one (see \Cref{sec:stats}).

\subsection{Sampling Definite Claims}\label{subsec:negative_sampling}

To ensure robust evaluation, we balance the dataset between definite and indefinite claims.
We use claims from applications whose office actions do not contain the term \enquote{112(b)}, i.e., applications whose claims are all definite.
This makes sure that all claims labelled as definite are indeed not rejected for indefiniteness; using claims from the same applications as the previously extracted indefinite claims could introduce noise if the parsing pipeline does not detect all rejections.
To also balance the number of applications in each class, we first compute the average number of indefinite claims per application included in the dataset.
We iterate through the definite applications in random order and sample the same number of claims until there are as many definite as indefinite claims.
While sampling claims from the applications, we round the number of sampled claims down or up depending on whether there are currently more or less claims per application than in the indefinite samples.

\subsection{Dataset Splits}

We randomly split the resulting dataset into train (60\%), test (30\%), and validation (10\%).
To avoid data leakage, we perform this split on the application-level, such that claims from the same application are always in the same split.

\subsection{Human Validation Study}\label{sec:human_eval}

To validate the quality of our extracted annotations, we manually inspect 50 randomly sampled claims from \DatasetName.
Among these claims, 24 are definite and 26 are indefinite, with a total of 27 reasons for indefiniteness (one indefinite claim has two reasons while the rest have one).
For each claim, we compare the rejection document with the extracted annotations and analyze the correctness of the binary label, the extracted reasoning texts, and the assigned categories.
We find that all binary labels are correct.
The extracted free-form reasoning is also correct in all samples except one, where it says \textit{\enquote{The underlined lacks antecedent basis}}, but this formatting is not translated into Markdown.
This validates the reliability of our automatically extracted binary labels and reasoning texts.
However, we find substantial noise in the assigned categories, with 19 out of 27 reasons having the correct one.
In six out of eight incorrect categories, none of the proposed categories would have been a good fit.
The LLM assigned the category \textit{\enquote{undefined term}}, instead of \textit{\enquote{other}} as instructed.
In the remaining two out of eight incorrect category assignments, the correct category would have been \textit{\enquote{antecedent basis}}, but the LLM assigned \textit{\enquote{undefined term}}.
In both cases, the term \textit{\enquote{antecedent basis}} was not mentioned explicitly, unlike in most reasoning texts of this category.

\subsection{Dataset Statistics}\label{sec:stats}

\Cref{tab:stats} shows dataset statistics over the splits.
In total, the dataset contains 14.5k claims from 3.7k patent applications, half of which are indefinite.
It is thus a balanced binary classification dataset, and a multi-label classification dataset considering the presence of each category as a binary label.
It consists of roughly 77\% dependent\footnote{A dependent claim is a claim that refers back to another claim, incorporating its features and narrowing its scope of protection.}
claims and 22\% independent claims.
The fraction of definite claims, the fraction of independent claims, and the distribution of rejection reasons are roughly the same in each split, i.e., there are no substantial distribution shifts.

\paragraph{Indefiniteness Categories.}
\Cref{tab:stats} also shows the distribution of indefiniteness categories.
Missing or ambiguous antecedent bases and unclear definitions dominate the dataset, constituting a combined 73\% of all indefiniteness reasons.
Relative terms and functional claiming are also commonly cited, each constituting about 9\% of indefiniteness reasons.
Contradicting limitations, exemplary phrasing and omission of essential elements or steps are each cited in less than 5\% of indefiniteness reasons.

\begin{figure}[t]
    \centering
    \input{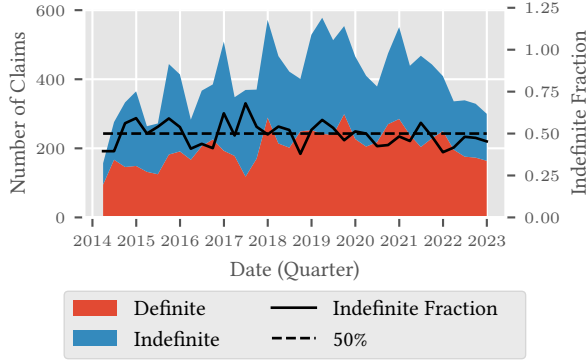}
    \caption{Definite vs.\ indefinite claims in our dataset over time. Left y-axis: stacked absolute values. Right y-axis fraction of indefinite claims. The fraction remains around 50\%, therefore models cannot determine definiteness by inferring the date.}
    \label{fig:definiteness_over_time}
\end{figure}

\paragraph{Characteristics of definite vs indefinite claims.}
If superficial characteristics differ between definite and indefinite claims, a trained classifier will likely use them as a shortcut.
We therefore report several such characteristics across the classes.
As shown in \Cref{fig:definiteness_over_time}, the fraction of indefinite claims remains around 50\% over time, i.e., models gain no advantage from the filing date, even if they infer it from the specific technology and terminology in the claim.
As shown in \Cref{tab:claim_characteristics}, the indefinite claims are more frequently independent claims than definite claims.
In other words, in our dataset, independent claims are more likely to be rejected due to indefiniteness than dependent claims.
A plausible reason is that independent claims are longer, i.e., they introduce more features that could be indefinite.
We consequently observe that indefinite claims are longer on average in terms of characters, words, and features.

\section{Definiteness Prediction}

In this section, we present two baseline approaches to predict indefiniteness: logistic regression and an LLM agent.
The available input is the claim in question and the accompanying patent description. 

\subsection{Logistic Regression}

We include logistic regression in our experiments because it is computationally efficient and interpretable, and because \citet{ashtorModelingPatentClarity2022} has shown it to achieve non-trivial performance predicting indefiniteness.
We use TF-IDF features and a number of handcrafted linguistic features, similar to \citet{ashtorModelingPatentClarity2022}.
The latter include the claim length, the claim length relative to the description length, readability metrics, trigger word flags, and a flag indicating whether the claim is independent. 
The full list is visualized in \Cref{fig:feature_importance}.
For all features sets, we train separate classifiers for the binary classification and the multi-label classification.

\subsection{LLM Agent}

LLMs have shown impressive performance on many tasks related to patents \cite{knappich2024pap2pat}, yet they have not been evaluated on definiteness prediction.
We implement a zero-shot LLM agent to identify the issues causing indefiniteness in the claims.
The prompt (see \Cref{sec:appendix_prediction_prompt}) includes instructions and the claim in question.
The agent is equipped with two tools that allow it to search for relevant information in the remaining document.
First, we add a tool that returns a claim given its claim number, allowing the agent to analyze parent claims, and other claims deemed relevant.
Second, we implement a TF-IDF search tool that allows the agent to retrieve paragraphs containing certain key words or phrases from the patent's description section.
We choose this tool-based approach because pasting the entire description quickly fills up the context window, while pre-selecting the relevant parts is restrictive and requires hand-crafting selection criteria and retrieval mechanisms.
Using these tools, the agent can flexibly retrieve whatever information it deems relevant.
The agent analyses the claim and performs tool calls in an interleaved fashion until it arrives at a final prediction.

Rather than generating the binary label directly (definite/indefinite), we instruct the LLM to use a verbalized expression of likelihood among a set of possible options ranging from \enquote{almost no chance} to \enquote{almost certain}.
Confidence scores, if reliable, should allow users to configure a level of strictness for issue detection.
Each expression is converted to a numerical probability according to the empirically determined human perception of probability expressions by \citet{fagen-ulmschneiderPerceptionProbabilityWords}.
We choose this approach because \citet{tian-etal-2023-just} and \citet{xiong2024can} show that asking the LLM for a confidence value is equally or more reliable than logit-based confidence estimation.

Lastly, we instruct the LLM to format the result as JSON according to a fixed schema.
The final output contains a prediction of the likelihood of the claim being rejected, and a list of potential reasons for the indefiniteness, where each reason contains a confidence score (using the same verbalized expressions as above), a free-form reasoning, one of the categories in \Cref{tab:categories}, and a list of claim recitations.
Thus, the output is in the same format as the structured representations extracted from the office actions, with the additional confidence scores.
This allows quantitative evaluation of the binary classification, multi-label classification, and the correctness of textual reasons.

\begin{table}[t]
    \centering
    \setlength{\tabcolsep}{4.1pt}
    \begin{tabular}{lrrrrrrr}
    \toprule
            & \multicolumn{3}{c}{Definite} & \multicolumn{3}{c}{Indefinite}                                                                                          \\
    \cmidrule(lr){2-4}\cmidrule(lr){5-7}
    \multicolumn{1}{r}{Independent} & \multicolumn{1}{c}{\yes + \no}                     & \multicolumn{1}{c}{\yes} & \multicolumn{1}{c}{\no} & \multicolumn{1}{c}{\yes + \no}   & \multicolumn{1}{c}{\yes} & \multicolumn{1}{c}{\no} \\
    \midrule
    \% Samples                      & 50.0\%                   & 7.8\%                    & 42.2\%                  & 50.0\% & 15.0\%                   & 35.0\%                  \\
    \# Parents                      & 1.2                      & 0.0                      & 1.4                     & 1.0    & 0.0                      & 1.5                     \\
    \# Characters                   & 415.4                    & 931.8                    & 319.4                   & 597.9  & 1080.8                   & 391.5                   \\
    \# Words                        & 64.9                     & 143.2                    & 50.3                    & 93.0   & 166.6                    & 61.5                    \\
    \# Features                     & 2.7                      & 6.4                      & 2.0                     & 3.6    & 6.8                      & 2.2                     \\
    \bottomrule
\end{tabular}
    \caption{Claim characteristics across definite and indefinite claims, further disaggregated over independent and dependent claims. The fraction of independent claims is larger among indefinite claims than among definite claims, and indefinite claims are longer than definite claims on average. The latter is largely, but not entirely, a side-effect of the former, since independent claims are naturally much longer than dependent claims.}
    \label{tab:claim_characteristics}
\end{table}

\section{Evaluation Metrics}\label{sec:metrics}

In this section, we propose a set of metrics to evaluate claim indefiniteness prediction models.
First, since the core task is binary classification, we use established metrics that measure how well the system can predict the binary label \enquote{definite} or \enquote{indefinite}.
While this binary label is already interesting by itself, practical applications demand a more fine-grained and explainable approach.
It is equally or even more important to know \textit{why} a claim is indefinite to give actionable feedback to drafters.
To that end, we implement an LLM-as-Judge \cite{zheng2023judge} approach that determines whether an examiner-cited reason and a model-cited reason point to the same essential issue in the claim.
Lastly, we evaluate the task as a multi-label classification, where the overlap in the categories of cited reasons also indicate whether a model decided to accept or reject a claim for the right reasons.

\begin{figure*}[th]
    \input{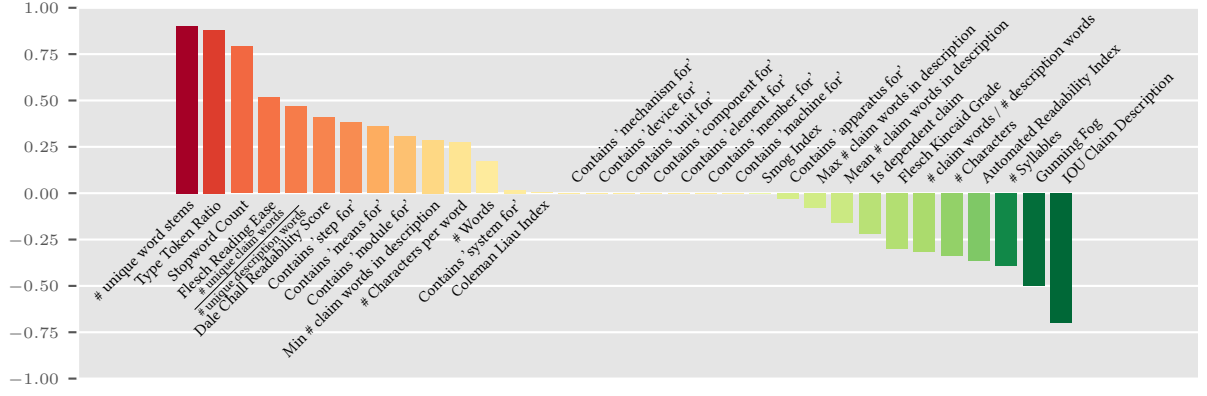}
    \caption{Feature importance scores of the logistic regression classifier with linguistic features. Positive values (red) indicate indefiniteness, negative values (green) indicate definiteness. The classifier performs similarly on train and test, i.e., the feature weights are not overfitted.}
    \label{fig:feature_importance}
\end{figure*}

\subsection{Binary Classification}
We evaluate binary classification performance for identifying indefinite claims using the following metrics:
\begin{list}{\textbullet}{\leftmargin=0.3cm \itemindent=0pt \labelwidth=0pt \labelsep=5pt \itemsep=0pt}
    \item \textbf{Precision}: The proportion of true positives (i.e., claims correctly predicted as indefinite) out of all claims predicted as indefinite by the model.
    \item \textbf{Recall}: The proportion of true positives out of all claims found indefinite by the examiner.
    \item \textbf{F1-score}: The harmonic mean of precision and recall, providing a balanced measure of both.
    \item \textbf{AUROC}: The area under the Receiver Operating Curve (ROC), which plots the true positive rate against the false positive rate for varying classification thresholds.
    \item \textbf{Accuracy}: The proportion of correctly classified claims (as either definite or indefinite) out of all claims.
\end{list}

\noindent These metrics provide a general understanding of the system's ability to distinguish between definite and indefinite claims.
For all models, we compute these metrics for a confidence threshold that balances the predictions, i.e., we first determine the threshold with which half of the claims from the validation set are predicted as indefinite.

\subsection{Pairwise Reasoning Judge}
To assess the quality of the model's reasoning for indefiniteness, we employ a reference-based LLM-as-Judge approach \cite{zheng2023judge}.
We prompt Gemma 3 27B to evaluate whether an examiner-cited reason and a model-cited reason point to the same essential issue in the claim on a scale from 1 (worst) to 5 (best) (we normalize the final scores to the range $[0, 100]$).
The few-shot prompt is attached in \Cref{sec:appendix_judge_prompt}.
To allow the model to analyze the ground truth and the predicted reason before settling on a grade, we first prompt it to find similarities and differences, then ask for a numerical grade in a follow-up message.
Following latest research in LLM-as-Judge systems \cite{wang2025improvingllmasajudgeinferencejudgment,yasunaga2024almaalignmentminimalannotation}, we use the probability-weighted mean as the similarity score of a reason-pair:
\[
    \text{norm}(s) = 100 \cdot \frac{s - 1}{4}\quad\quad\text{sim} = \text{norm}\left(\frac{\sum_{i=1}^5 i \cdot p_i}{\sum_{i=1}^5 p_i}\right)
\]
where $i$ is a grade and $p_i$ is the LLM-determined probability of that grade (i.e., the exponential of the logit of the token corresponding to the grade).

We compute this similarity for every pair of examiner-cited reason and model-cited reason.
Hence, for a sample with $n$ examiner-cited reasons and $m$ model-cited reasons, we obtain a similarity matrix $S \in \mathbb{R}^{n \times m}$.
From this matrix, we compute precision, recall and F1, each in a thresholded and a soft variant.
In the thresholded variant, we consider a model-cited reason correct if there is an examiner-cited reason with which it has a similarity score of 75 (corresponding to \textit{\enquote{The reasons are closely related and largely address the same issue in the claim}}) and above, and vice versa.
\begin{align*}
    \text{P}_{\ge75} & = \frac{1}{m} \sum_{j=1}^{m} \max_{i} \mathds{1}_{S_{ij} \ge 75} \\
    \text{R}_{\ge75} & = \frac{1}{n} \sum_{i=1}^{n} \max_{j} \mathds{1}_{S_{ij} \ge 75}
\end{align*}
We report the macro average (i.e., the average of all average per-claim scores) and the micro average (i.e., the average of all per-reason scores).
For the soft variant, we do not set a fixed threshold but directly average over the maximum similarity in the respective dimension.
\[
    \text{P} = \frac{1}{m} \sum_{j=1}^{m} \max_{i} S_{ij}\quad\quad\text{R} = \frac{1}{n} \sum_{i=1}^{n} \max_{j} S_{ij}
\]
In both cases, F1 is computed as harmonic mean of P and R.

\subsection{Multi-Label Classification}
We also assess the system as a multi-label classification task, where each label serves as a binary indicator denoting the presence or absence of a specific category within the list of indefiniteness reasons.
A high score means that the model has classified a claim for reasons belonging to the same or overlapping indefiniteness categories as the examiner, and thus also represents a kind of explanation for the binary classification.
We report the micro and macro average of the category-wise F1 scores.
As with the binary classification, we balance the predictions using the confidence threshold per category under which the fraction of positive predictions matches the distribution in the validation set.

\section{Results}

\Cref{tab:results} shows the classification results.
The best overall model achieves an AUROC score of 60.3.
Thus, its predictions are considerably better than random, but there is substantial room for improvement.
LLM agents do not outperform the logistic regression baseline in terms of binary classification performance.
Despite this, we find that the agents correctly identify many of the reasons for indefiniteness that were also noted by the examiner.
In the following, we further analyze the classification performance, classifier calibration, and the LLM-as-Judge results.

\begin{table}[t]
    \centering
    \setlength{\tabcolsep}{2.1pt}
    \footnotesize
    \begin{tabular}{lcccccccccccccccc}
    \toprule
    \multirow{2}{*}{Model} & \multirow{2}{*}{\% indef} & \multicolumn{5}{c}{Binary} & \multicolumn{2}{c}{Multi-Label} \\
    \cmidrule(lr){3-7} \cmidrule(lr){8-9}
    & & P & R & F1 & AUC & Acc & \makecell{Macro\\F1} & \makecell{Micro\\F1}  \\
    \midrule

    Random &
    50.4\% &
    50.9 &
    51.4 &
    51.1 &
    51.3 &
    50.9 &
    6.6 &
    11.9 & \\

    \multicolumn{9}{c}{\textit{Logistic Regression}}\rule{0pt}{3ex}\\

    TF-IDF &
    46.2\% &
    53.2 &
    49.3 &
    51.2 &
    54.8 &
    53.0 &
    9.3 &
    18.5 & \\

    Ling. Features &
    49.2\% &
    57.2 &
    56.3 &
    56.8 &
    59.4 &
    57.2 &
    11.8 &
    21.1 & \\ 

    All Features &
    49.2\% &
    56.8 &
    55.9 &
    56.3 &
    59.5 &
    56.7 &
    12.1 &
    22.0 & \\  

    \multicolumn{9}{c}{\textit{LLM Agents}}\rule{0pt}{3ex} \\

    Qwen 2.5 32B &
    45.9\% &
    54.7 &
    50.3 &
    52.4 &
    55.2 &
    54.4 &
    13.7 &
    22.8 & \\  

    \hspace{1em} + Log. Reg. &
    48.3\% &
    55.8 &
    54.0 &
    54.9 &
    58.8 &
    55.7 &
    13.7 &
    22.7 & \\ 

    Qwen 2.5 72B &
    33.7\% &
    57.8 &
    39.0 &
    46.6 &
    56.8 &
    55.3 &
    13.5 &
    21.7 & \\ 

    \hspace{1em} + Log. Reg. &
    48.9\% &
    \textbf{59.3} &
    \textbf{58.2} &
    \textbf{58.8} &
    \textbf{60.3} &
    \textbf{59.2} &
    \textbf{14.1} &
    \textbf{22.9} & \\ 

    \bottomrule
\end{tabular}
    \caption{
        Classification Results. 
        The best value per column is bolded. All predictions use the prediction threshold that balances the label distribution on the validation set. P=Precision, R=Recall, Acc=Accuracy, AUC=AUROC. The random baseline samples from a uniform distribution.
    }
    \label{tab:results}
\end{table}

\begin{figure}[t]
    \input{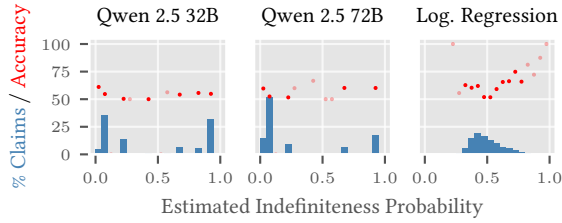}
    \caption{
        Distributions of claim-level indefiniteness probability estimates for different binary classifiers and the relation of estimated probability and accuracy. 
        The accuracy per probability bin is plotted as red dots. 
        For bins with less than 1\% of samples, the dot is faded. 
        The logistic regression model shown here uses the full feature set. 
    }
    \label{fig:confidence_dist}
\end{figure}

\paragraph{Logistic Regression.}
Among the three proposed variants of logistic regression, the one with linguistic features and the one with all features perform comparably, while using only TF-IDF features performs substantially worse.
This indicates that the handcrafted features are indeed helpful to differentiate definite and indefinite claims, and that TF-IDF features provide only limited additional information.

We show the feature importance values in \Cref{fig:feature_importance}.
The most important feature indicating definiteness is the IOU of words between claim and description, i.e., samples with a high word overlap between claim and description are less likely to be rejected for indefiniteness.
The highest-weighted features indicating indefiniteness are the number of unique word stems, the type token ratio and the number of stopwords.
Among the binary features showing whether a trigger word is contained in the claim, \enquote{step for} is the most indicative of indefiniteness.
The flag whether the claim is independent indicates definiteness, which is in line with our dataset statistics, albeit with moderate importance.
Many of the text complexity and readability metrics have high importance scores.
A higher text complexity is associated with definiteness (with Gunning Fog, Automated Readability Index, and Flesch Kincaid Grade having a negative feature weight) and better readability is associated with indefiniteness (with Flesch Reading Ease and Dale Chall Readability Score having a positive feature weight).
While this might seem counterintuitive when relating definiteness to clarity or understandability, it makes sense when considering definiteness as the absence of ambiguity, as removing ambiguities often adds complexity through additional specification.

\begin{table}[t]
    \setlength{\tabcolsep}{2.7pt}
    \begin{tabular}{cccccccc}
    \toprule
    & &
    \multicolumn{3}{c}{Soft} &
    \multicolumn{3}{c}{Thresholded}
    \\
    \cmidrule(lr){3-5} \cmidrule(lr){6-8}
    &
    &
    $P$ &
    $R$ &
    $F1$ &
    $P_{\ge75}$ &
    $R_{\ge75}$ &
    $F1_{\ge75}$
    \\
    \midrule
    \multirow{2}{*}{Qwen 2.5 32B} & 
    macro & 
    \textbf{14.6} & 
    \textbf{43.9} & 
    \textbf{21.9} & 
    \textbf{11.2} & 
    \textbf{17.9} & 
    \textbf{13.8} \\
    &
    micro & 
    \textbf{10.8} & 
    \textbf{43.6} & 
    \textbf{17.3} & 
    \textbf{7.7} & 
    \textbf{35.4} & 
    \textbf{12.7} \\
    \cmidrule{1-8}
    \multirow{2}{*}{Qwen 2.5 72B} & 
    macro & 
    9.8 & 
    33.5 & 
    15.1 & 
    7.4 & 
    12.3 & 
    9.2 \\
    & 
    micro & 
    6.1 & 
    33.6 & 
    10.3 & 
    4.0 & 
    25.0 & 
    6.9 \\
    \bottomrule
\end{tabular}

    \caption{LLM-as-Judge results using Gemma 3 27B as judge. P=Precision, R=Recall. The thresholded variant considers reasons with scores above 75\% to be correct.}
    \label{tab:judge}
\end{table}

\begin{figure}[t]
    \centering
    \input{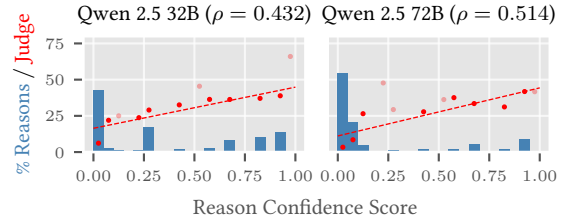}
    \caption{
        Distributions of reason-level confidence scores and the relation of confidence and correctness according to LLM-as-Judge. 
        The reported judge scores are the soft micro-averaged precision scores, and thus represent the average max similarity per model-cited reason. 
        For bins with less than 1\% of samples, the dot is faded. 
        The dashed line is a linear regression that ignores the faded points.
        $\rho$ corresponds to the pearson correlation between confidence and judge score.
    }
    \label{fig:confidence_dist_judge}
\end{figure}

\paragraph{LLM Agent.}
The LLM agents using Qwen 2.5 32B and 72B both make use of the provided tools, each making at least one tool call for 98\% of the samples.
Notably, the larger model makes more tool calls (3.3 calls per sample on average) than the smaller model (2.3 calls per sample on average).
This indicates that the larger model is more thorough in searching for relevant information in the document, which could be a reason for the slightly higher AUROC and accuracy.
Recall and F1 are notably worse for the larger model, caused by the imbalance of its predictions.
Both metrics favor classifiers that frequently predict indefiniteness: F1 is per definition $\frac{2}{3}$ when predicting indefiniteness for every sample.
The distribution of confidence scores produced by the 72B model makes it impossible to find a threshold that moves the fraction of indefinite predictions close to 50\%.
We show the distribution of confidence scores for both LLM agents and the logistic regression in \Cref{fig:confidence_dist}.
The LLM agents are typically highly confident in either extreme, whereas the logistic regression produces a smoother distribution with a peak in the middle.
In addition, as visible in \Cref{fig:confidence_dist}, there is no clear relationship between confidence and accuracy in the LLMs' binary predictions.
This is unlike the predictions of the logistic regression model, where the accuracy drops notably around the classification threshold.
Overall, the results indicate that the LLMs confidence estimates are not well-calibrated, and that better calibration could enhance the LLMs' classification performance.

\paragraph{Multi-Label Classification.}
LLM agents and logistic regression achieve only moderate F1 scores.
All models achieve a much better micro F1 than macro F1, indicating that the performance varies between categories.
All models perform much better on the well-represented categories than on rare ones.
Given the substantial noise in the labels for the category \enquote{\textit{undefined term}} (see \Cref{sec:human_eval}), additional investigations are necessary to draw further conclusions.

\paragraph{Ensemble.}
We achieve the best overall performance using an ensemble between Qwen 2.5 72B and the logistic regression.
To create the ensemble, we average the predicted probability of indefiniteness for every sample.
This leads to moderate improvements, indicating that the predictions are complementary to some degree.

\begin{table*}[t]
    \footnotesize
    \begin{tabular}{cp{.6\linewidth}p{.21\linewidth}c}
        \toprule
         &
        Examiner & 
        Qwen 2.5 32B &
        \makecell{Judge\\Similarity}
        \\
        \midrule

        1 &
        The term \highlight{yellow}{'right before'} in claim 9 is a \highlight{green}{relative term} which renders the claim indefinite. The term 'right before' is not defined by the claim, the specification does not provide a \highlight{highlightblue}{standard for ascertaining the requisite degree}, and one of ordinary skill in the art would not be reasonably apprised of the scope of the invention. The amount of time necessary for an input to constitute being obtained 'right before' the first input is unclear. Furthermore and for example, a person of ordinary skill in the art may consider an arbitrary number of previous inputs, such as three, to constitute inputs obtained 'right before' a first input. &
        The term \highlight{yellow}{'right before'} is a \highlight{green}{relative term} that lacks a clear, \highlight{highlightblue}{objective standard for} \highlight{highlightblue}{determining} the exact input used to determine the first domain. &
        100\% 
        \\
        \cmidrule[0.001pt]{1-4}

        2 &
        The claims recite the limitation \highlight{yellow}{'ranking the one or more groups'} There is \highlight{green}{insufficient antecedent basis} for this limitation in the claim, it appears likely that \highlight{highlightblue}{this claim should instead depend from claim 2}. &
        The term \highlight{yellow}{'one or more groups'} in claim 3 \highlight{green}{lacks antecedent} \highlight{green}{basis} as it is not introduced or defined in claim 1. &
        75\%
        \\
        \cmidrule[0.001pt]{1-4}

        3 &
        As per Claim 10, '\highlight{green}{each} \highlight{yellow}{highlight} in the second pane in the second interface' \highlight{highlightblue}{lacks} \highlight{highlightblue}{antecedent basis} to the extent that \highlight{green}{'each' implies that there is more than one highlight} in the second pane. Claim 9 only recites where a single highlight is dropped into the second pane and Claim 1 recites where selected ones of the highlights are associated with the second pane of the second interface in a sequence (associating does not imply putting the selected highlights into the second pane of the second interface, and associating in a sequence can be where a sequence in the second interface includes elements referencing to the highlights themselves without including the highlights themselves). &
        The term \highlight{yellow}{'highlight panel'} is not clearly defined in the claims or the description, leading to uncertainty about its scope. &
        25\%
        \\       
        \bottomrule
    \end{tabular}
    \caption{Examples of examiner-cited reasons and best matching model-cited reasons for indefiniteness. Important parts are highlighted with matching colors between examiner and model if applicable.}
    \label{tab:discussion_examples}
\end{table*}

\paragraph{Judge Evaluation.}
Our LLM-as-Judge evaluation provides further insight into LLM agents' performance by directly analyzing the quality of their identified reasons for indefiniteness.
\Cref{tab:judge} shows the results for Qwen 2.5 32B and 72B.
Interestingly, the smaller model clearly outperforms the larger one in all judge metrics, despite exhibiting slightly lower binary classification performance.
There seems to be a disconnect between the models' ability to identify potential reasons for indefiniteness and its ability to accurately decide whether a claim is indefinite.
The 32B model is able to correctly identify 35.4\% of the examiner-cited reasons (micro $R_{\ge75}$, 25.0\% for 72B).
Both models list many reasons, even if they consider them unlikely to cause indefiniteness, causing the precision to be lower than the recall.

\Cref{fig:confidence_dist_judge} also shows the relation between reason-level confidence and the soft micro-averaged precision.
Both models exhibit a positive trend; as their confidence in a reason increases, the likelihood of the examiner citing the same or a similar reason also increases.
The Pearson correlation between confidence and precision is positive and moderate ($\rho=0.432$ for 32B and $\rho=0.514$ for 72B).
That is, the reason-level confidence appears to be better calibrated than the claim-level confidence.
Future work should further investigate the relation between these two sub-tasks and develop methods to bridge this disconnect.

\section{Discussion}

We here compare example reasons for indefiniteness generated by Qwen 2.5 32B with those written by the human examiner.
\Cref{tab:discussion_examples} lists three examples.
In Example 1, the LLM agent correctly identified the reason for indefiniteness, albeit with a less verbose explanation.
In Example 2, the LLM agent also found the correct underlying issue. 
However, its explanation is not equally helpful as the one provided by the human examiner, which additionally suggests that the claims might accidentally depend from the wrong base claim.
In Example 3, the model-generated reason points to similar problematic phrases, but without identifying the underlying issue. 
The LLM-as-Judge correctly identifies that there is little substantial overlap and assigns a lower score of 25\%.
Generally, the model-generated reasons seem to be less detailed, and often lack the deep analysis and reasoning performed by human examiners.

While our dataset creation process is domain-agnostic, our study focuses on patents from the NLP domain.
Therefore, the claim language and terminology is relatively homogeneous.
Other domains might differ in terms of terminology, distribution of claim types and categories of indefiniteness, and legal conventions.
Future work should generalize our study design and verify that our findings transfer to other domains.

\section{Conclusion and Outlook}

In this work, we tackle the task of automatic patent definiteness examination.
We present \DatasetName, the first publicly available dataset for this task to enable reproducible experiments with different examination models.
We conduct first experiments on \DatasetName using logistic regression and LLMs.
All models are able to perform better than random, but there is substantial room for improvement.
LLMs are able to identify many of the reasons for indefiniteness also cited by the human examiner, yet their binary classification performance fails to clearly outperform logistic regression with hand-crafted features.
We show that poor calibration is one of the issues of current LLM-based setups.  
Promising directions for future research include better calibration methods, fine-tuning on task-specific data using supervised and/or reinforcement learning, and using the rejection full-text for in-depth evaluation of other patent requirements like novelty and non-obviousness.

\begin{acknowledgments}
We would like to thank the patent attorneys Philipp Mangold and Charlotte Hellmann for insightful discussions and helpful pointers.
\end{acknowledgments}

\section*{Declaration on Generative AI}

During the preparation of this work, the authors used Gemini 2.0 Flash in order to\footnote{\url{https://ceur-ws.org/GenAI/Taxonomy.html}}: grammar and spelling check, improve writing style, paraphrase and reword.
After using these tool(s)/service(s), the authors reviewed and edited the content as needed and take full responsibility for the publication's content.

\bibliography{bibliography}

\appendix

\begin{figure*}[t]
    \section{Office Action Parsing Prompt}\label{sec:appendix_parsing_prompt}
    \begin{lstlisting}[basicstyle=\ttfamily\footnotesize, xleftmargin=.4cm, caption=Prompt used to parse office action into the described JSON schema. \texttt{schema} and \texttt{rejection\_categories} are replaced with their respective values.]
### TASK

Your task is to extract data related to claim rejections under 35 U.S.C. 112(b) or pre-AIA 35 U.S.C. 112, second paragraph (indefiniteness).

You will be provided with a snippet of text from a USPTO Office Action. You will identify claim rejections based on indefiniteness (35 U.S.C. 112(b) or pre-AIA 35 U.S.C. 112, second paragraph) and represent them in a JSON format.

### OUTPUT FORMAT

Strictly adhere to the following JSON schema and reply with nothing but the JSON:

```json
{schema}
```

### GUIDELINES

Here is a detailed guide on how to extract the information:

1. Identify the section that describes rejections regarding 35 U.S.C. 112(b).
2. Identify which claims were rejected based on 35 U.S.C. 112(b). Note this in the "rejectedClaims" field. You can either list individual claims as integers or inclusive claim ranges (e.g., 1-3 equals 1,2,3) as strings (ranges are preferred if applicable).
3. Next, extract the individual reasons for rejection based on 35 U.S.C. 112(b) and list them under "rejectionReasons". For each reason, determine the following properties:

    3.1 "reasonText": The full text describing the rejection. Copy this verbatim. Do not include statements regarding interpretation ("For purpose of examination ...").

    3.2 "reasonContext": The reasoning context. Use this if the reasonText reads in the lines of "Claim X is rejected for similar reasons as claim Y", and add the reason of claim Y here. Make sure enough context is given to make the rejection undestandable on its own. Leave this field empty if the reasonText does not refer to another reason.

    3.3 "claims": The mentioned claims. You can either list individual claims as integers or inclusive claim ranges (e.g., 1-3 equals 1,2,3) as strings (ranges are preferred if applicable).

    3.4 "reasonCategory": The category of 112(b) rejection. Choose the most appropriate among the following categories. You can use only those exact values!
{rejection_categories}

    3.5 "claimRecitations": The parts of the claim(s) in question that the examiner cited as causing the indefiniteness. They are often, but not always, included in quotes. Copy verbatim. Every entry here should be a substring of the claim text, i.e., do NOT include the reasoning around recitations. Include EVERY SINGLE recitation in the reasonText, even if they are repeated or overlapping!.

Some further notes:

*   **Focus solely on 35 U.S.C. 112(b) (or pre-AIA 35 U.S.C. 112, second paragraph) rejections.**  Ignore other types of rejections (e.g., 112(a), 102, 103).
*   **Extract information verbatim. Do not paraphrase or summarize any text. Do not fix spelling or grammar errors.** Ignore all newlines and consistently use single quotes.
*   **Do not mix multiple reasons in a single "rejectionReasons" entry.** Extract rejection reasons as fine-grained as possible.
*   If unsure about the `reasonCategory`, use `other`.
*   Output only valid JSON. Do *not* include any explanatory text before or after the JSON. If there are no 35 U.S.C. 112(b) claim rejections to extract from the provided text, leave "rejectionReasons" empty.

From now on, I will only send an office action document and you will respond with the corresponding JSON representation. Do you understand?        
    \end{lstlisting}
\end{figure*}

\begin{figure*}[t]
    \section{Indefiniteness Examination Prompt}\label{sec:appendix_prediction_prompt}
    \begin{lstlisting}[basicstyle=\ttfamily\footnotesize, xleftmargin=.4cm, caption=Prompt used to examine a given claim with respect to indefiniteness. \texttt{indefiniteness\_categories} and \texttt{claim} are replaced with their respective values.]
Examine this patent claim with respect to definiteness.

### Guidelines

- Carefully dissect the claim and its features and search for common patterns that cause indefiniteness.
- Think step by step and reason about potential issues! You can correct yourself at any point in time, only the final verdict counts.
- Be very thorough and list all potential issues you can find!
- Ultimately, estimate the likelihood of the claim begin rejected due to indefiniteness. Note that a single issue renders the entire claim indefinite. 
- Completely ignore all other aspects like novelty or non-obviousness, focus entirely on indefiniteness.

### Categories of Indefiniteness

{indefiniteness_categories}

### Tools

- You are given access to the remaining patent application using simple tools.
- You can get other claims (e.g. parent claims) using `get_claim` and retrieve relevant description paragraphs using `search_description`.
- If you want to search for multiple terms, invoke `search_description` separately for each term. This will yield much more specific results than using a combined query.
- You can perform as many tool calls as you like; make sure you have all the necessary information.

### Claim

{claim}   
    \end{lstlisting}
\end{figure*}

\begin{figure*}[t]
    \section{LLM-as-Judge Prompt}\label{sec:appendix_judge_prompt}
    \begin{lstlisting}[basicstyle=\ttfamily\footnotesize, xleftmargin=.4cm, caption=Prompt used to evaluate the similarity of two reasons for indefiniteness. \texttt{reason\_1} and \texttt{reason\_2} are replaced with their respective values.]
<instruction>
You will evaluate the performance of an AI system used to identify indefiniteness issues in patent claims. You will be given 2 short text snippets that mention an issue in a claim that renders it indefinite, one written by a human examiner, one written by the AI. Your task is to determine whether the two text snippets refer to the same issue or not. Ignore differences in phrasing and evaluate only whether they pinpoint the same issue. You can always assume that both text snippets talk about the same claim. Use the scale shown below.
</instruction>

<scale>
1: The reasons are completely unrelated and address different concerns.
2: The reasons address distinct aspects of the claim with minimal overlap.
3: The reasons overlap in some areas but also have notable differences.
4: The reasons are closely related and largely address the same issue in the claim.
5: The reasons are essentially identical and address the same issue in the claim.
</scale>

<examples>
    <example>
        <text-1>
The phrase 'like' (stated in all the claims) renders the claim(s) indefinite because the claim(s) include(s) elements not actually disclosed (those encompassed by 'like'), thereby rendering the scope of the claim(s) unascertainable. See MPEP 2173.05(d).
        </text-1>
        <text-2>
The claim does not specify how these 'component variables' are utilized or how they relate to the method described in Claim 13, which is critical to understanding the claim's scope.
        </text-2>
        <score>
        1
        </score>
    </example>
    <example>
        <text-1>
The phrase 'like' (stated in all the claims) renders the claim(s) indefinite because the claim(s) include(s) elements not actually disclosed (those encompassed by 'like'), thereby rendering the scope of the claim(s) unascertainable. See MPEP 2173.05(d).
        </text-1>
        <text-2>
The use of 'like' in 'example component variables of solution automation & interface analysis (like 'solution automation workflow variables')' suggests that these might be examples rather than exhaustive lists, leading to potential ambiguity.
        </text-2>
        <score>
        5
        </score>
    </example>
    <example>
        <text-1>
Claim 1 recites the limitation 'a corresponding link instruction' (line 2) and 'at least one link instruction' (line 12). It would be unclear to one having ordinary skill in the art whether the above limitations are intended to be identical to, common to, or distinct from one another.
        </text-1>
        <text-2>
The term 'link instruction' is not clearly defined in the specification, leading to ambiguity in the scope of the claim.
        </text-2>
        <score>
        4
        </score>
    </example>
</examples>

Apply this scheme, as shown in the examples, to the text snippets shown below.

<text-1>
{reason_1}
</text-1>

<text-2>
{reason_2}
</text-2>

Before you settle on a score, summarize the similarities and differences between the two reasons for indefiniteness. 
    \end{lstlisting}
\end{figure*}

\end{document}